\documentclass{article}

\usepackage{PRIMEarxiv}

\usepackage[utf8]{inputenc} 
\usepackage[T1]{fontenc}    
\usepackage[hidelinks]{hyperref}
\usepackage{url}            
\usepackage{booktabs}       
\usepackage{amsfonts}       
\usepackage{nicefrac}       
\usepackage{microtype}      
\usepackage{lipsum}
\usepackage{graphicx}
\usepackage{placeins}
\graphicspath{{media/}}     

\title{Prompt-Driven Image Analysis with Multimodal Generative AI: Detection, Segmentation, Inpainting, and Interpretation
}

\author{
  Kaleem Ahmad \\
  Independent Researcher \\
\texttt{imkaleemahmad@gmail.com} \\
}

\begin{document}
\maketitle

\begin{abstract}
Prompt-driven image analysis turns a single natural-language instruction into a series of steps: locate, segment, edit, and describe. We present a practical case study of a unified pipeline that integrates open-vocabulary detection, promptable segmentation, text-conditioned inpainting, and vision-language description into a single workflow. The system operates end to end from a single prompt, retains intermediate artifacts for transparent debugging (such as detections, masks, overlays, edited images, and before and after composites), and offers the same functionality through an interactive UI and a scriptable CLI for consistent, reproducible runs. We highlight integration choices that decrease brittleness, including threshold sweeps, mask inspection with light morphology, and resource-aware defaults. In a small single-word prompt segment, detection and segmentation produced usable masks in over 90\% of cases with an accuracy above 85\% based on our criteria (Section~\ref{sec:mini_quant}). On a high-end GPU, inpainting makes up 60 to 75\% of total runtime under typical guidance and sampling settings, which highlights the need for careful tuning. The study provides implementation-guided advice on thresholds, mask tightness, and diffusion parameters, and details version pinning, artifact logging, and seed control to support replay. Our contribution is a transparent, dependable pattern for assembling modern vision and multimodal models behind a single prompt, with clear guardrails and operational practices that improve reliability in object replacement, scene augmentation, and removal.
\end{abstract}


\section{Introduction}
Prompt-driven image analysis and editing aim to convert a single natural-language instruction into a series of visual actions, such as locate, segment, edit, and explain. This allows non-experts to work with complex scenes without manual masks or specialized tools. The essential components are now broadly accessible: open-vocabulary detectors, promptable segmentation, diffusion-based inpainting, and instruction-tuned vision-language models \cite{liu2023groundingdino,kirillov2023segmentanything,rombach2022latentdiffusion,liu2023visualinstruction}. Yet making these components function as a reliable, unified system remains difficult. Early errors cascade into later stages, prompt ambiguity destabilizes grounding, and reproducibility is hampered by heavyweight models, external APIs, and changing dependencies \cite{sculley2015hidden}. What is missing is an implementation-aligned plan for how to assemble these parts into a transparent, auditable pipeline that can be controlled end-to-end by a single prompt.

\begin{figure}[htbp]
  \centering

  \includegraphics[width=0.9\linewidth]{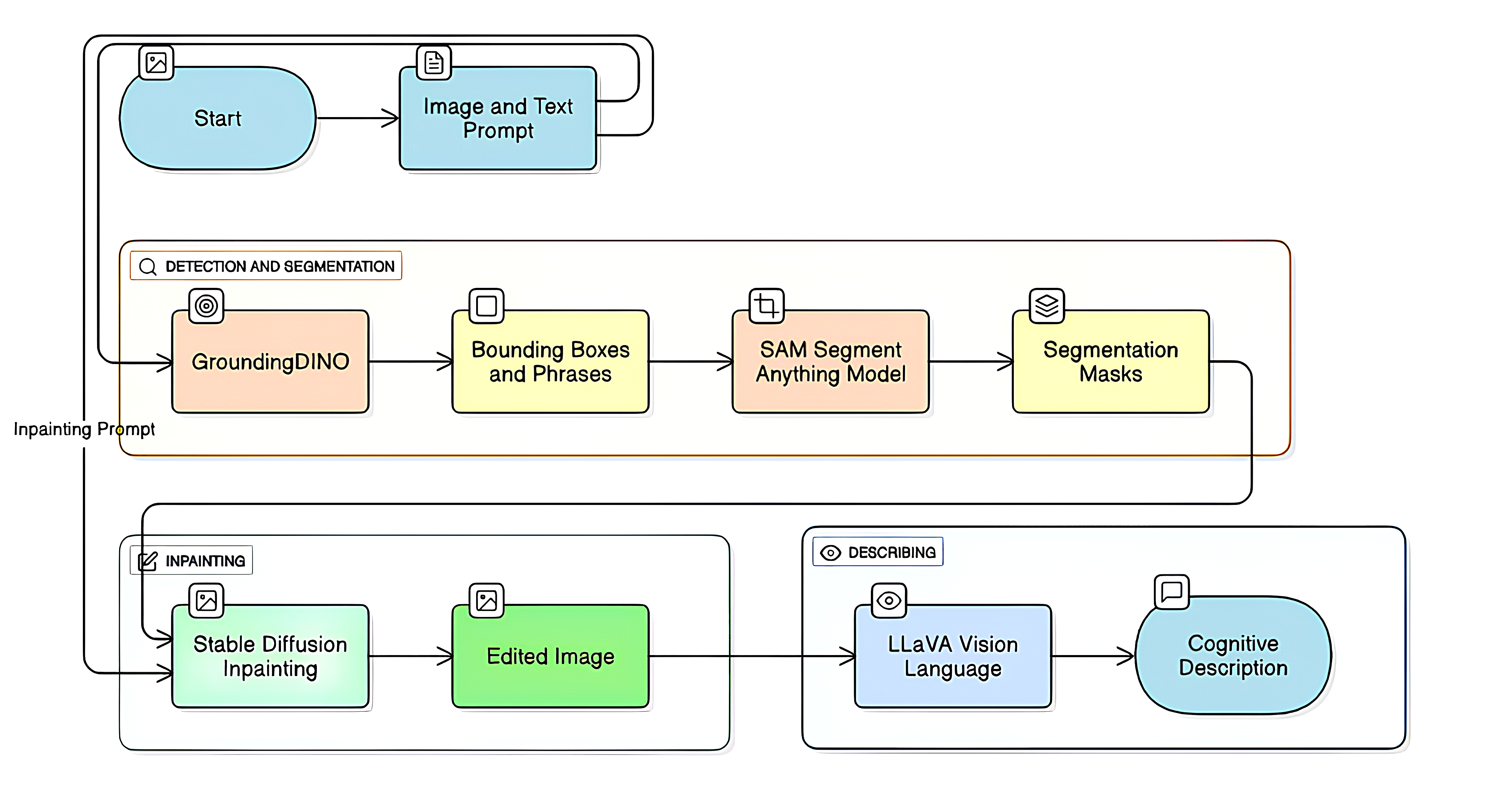}
  
  \caption{Pipeline overview of the four-stage prompt-driven system that integrates detection, segmentation, inpainting, and description. The diagram shows data flow between modules and highlights intermediate artifacts (bounding boxes, segmentation masks, edited images) that support stepwise validation and debugging.}
  \label{fig:system_overview}
\end{figure}

This capability is required by specific use cases. Creative professionals want to replace or enhance objects while maintaining lighting and perspective; product and interior designers need quick prototyping that respects scene layout; educators and communicators look for tools that both modify images and explain results in simple language. While previous work shows the components and even partial assemblies like Grounded SAM \cite{ren2024groundedsam}, practical guidance on robust composition such as thresholds, guardrails, artifact logging, and UI/CLI parity remains limited outside controlled benchmarks. \cite{liu2023groundingdino,kirillov2023segmentanything,rombach2022latentdiffusion,liu2023visualinstruction}. Figure~\ref{fig:system_overview} illustrates our end-to-end data flow and the persisted artifacts that support step-by-step validation. We introduce our default thresholds and diffusion guidance options here, with more details provided in Section~\ref{sec:method}.

We present a practical case study of a unified, single-prompt pipeline, which we call Locate–Segment–Inpaint–Describe (LSID). LSID integrates text-grounded detection (GroundingDINO), promptable segmentation (SAM), text-conditioned inpainting (latent diffusion), and multimodal interpretation (LLaVA) into a single workflow with artifact-level transparency. The system retains intermediate artifacts at each stage and provides access to the same backend through an interactive Gradio UI and a scriptable CLI for reproducible runs \cite{liu2023groundingdino,kirillov2023segmentanything,rombach2022latentdiffusion,liu2023visualinstruction}.

We organize the study around four practical questions:
\begin{itemize}
  \item \textbf{RQ1. Integration.} How can we integrate detector, segmenter, diffusion editor, and vision-language describer behind a single prompt while keeping the data flow transparent?
  \item \textbf{RQ2. Robustness.} Which guardrails (threshold sweeps, mask inspection/post-processing, prompt refinement) most effectively reduce compounding errors between stages?
  \item \textbf{RQ3. Operations.} What are the main resource and latency factors, and how should users adjust guidance and steps for reliable edits on common hardware?
  \item \textbf{RQ4. Reproducibility.} Which practices, such as version pinning, artifact logging, and secrets management, are enough to reproduce qualitative outcomes over time?
\end{itemize}

Our responses are based on empirical data and implementation considerations. Qualitatively, we document patterns for object replacement, scene enhancement, and removal, and we analyze failure cascades due to ambiguous language, weak detections, and mask leakage. Quantitatively, a small single-word prompt segment shows that GroundingDINO+SAM produce usable masks in over 90\% of cases with accuracy above 85\% according to our criteria (Section~\ref{sec:mini_quant}). Operationally, on a high-end GPU, inpainting makes up about 60–75\% of the total runtime under typical guidance and sampling settings, highlighting the importance of resource-aware configuration.

Our contributions are specific and verifiable:
\begin{itemize}
  \item \textbf{Implementation-aligned integration.} A single-prompt interface and scriptable pipeline that unify detection, segmentation, inpainting, and interpretation.
  \item \textbf{Transparent data flow.} Persisted artifacts at each stage serialized detections, annotated visualization, binary mask, color overlay, edited image, and before-and-after composite for stepwise validation and failure localization.
  \item \textbf{Robustness analysis and guardrails.} Practical strategies to reduce ambiguity and error spread, such as threshold sweeps, mask overlay inspections, light post-processing, and prompt refinement protocols.
  \item \textbf{Resource profile and configuration guidance.} Empirical latency analysis and guidance on diffusion steps, guidance scales, and SAM variant selection for reliability and cost.
  \item \textbf{Reproducibility practices.} Version pinning, artifact logging, and secrets management for external APIs to ensure accurate re-runs.
  \item \textbf{Mini-quantitative slice.} An indicative single-word prompt to complement the qualitative analysis (Section~\ref{sec:mini_quant}).
\end{itemize}

Scope and impact. We utilize off-the-shelf models without introducing new architectures. The value of this case study lies in practical integration patterns, clear reporting, and empirically supported guidance for reliable prompt-driven editors. The rest of the paper reviews related work, outlines the method and implementation, presents a qualitative case study with some quantitative data, and concludes with discussion, limitations, ethics, and future directions.

\FloatBarrier

\section{Related Work}
Prompt-driven image analysis combines open-vocabulary localization, promptable segmentation, diffusion-based editing, and multimodal understanding. Our focus is on composition and reliability rather than developing new architectures. We rely on widely used baselines and integrated systems, and we prioritize artifact-level transparency and reproducibility.

Open-vocabulary detection and cross-modal alignment. CLIP established transferable vision-language representations that enable text-to-region grounding 
\cite{radford2021clip}. GroundingDINO extends this approach with text-grounded detection for open-set scenarios, producing bounding boxes conditioned on phrases 
\cite{liu2023groundingdino}. Related pretraining strategies in the vision-language domain, such as BLIP and BLIP-2, improve captioning and instruction-following quality 
\cite{li2022blip,li2023blip2}. In our setup, GroundingDINO transforms user phrases into candidate regions that seed downstream segmentation.

Promptable segmentation and referring segmentation. The Segment Anything Model provides high-quality masks from prompts such as boxes and points \cite{kirillov2023segmentanything}. General-purpose segmentation frameworks, including Mask2Former and X-Decoder, move toward universal pixel-level reasoning across tasks \cite{cheng2022mask2former,xdecoder2023}. Referring image segmentation connects language spans to pixel regions \cite{cris2022,lavt2022,restr2022}, which complements our use of SAM to refine text-grounded boxes. Feature adapters like CLIP-Adapter and Tip-Adapter enhance alignment without full fine-tuning \cite{gao2021clipadapter,zhang2022tipadapter}.

Diffusion-based inpainting and control. Latent Diffusion underpins modern image editing and inpainting \cite{rombach2022latentdiffusion}. Methods such as RePaint and SDEdit explore sampling and noise-injection strategies for editing \cite{lugmayr2022repaint,meng2021sdedit}. Controllability advances include Prompt-to-Prompt, InstructPix2Pix, and ControlNet \cite{hertz2022prompt2prompt,brooks2023instructpix2pix,zhang2023controlnet}. Identity and color fidelity can be preserved via Paint-by-Example, Blended Diffusion, and Palette \cite{yang2022paintbyexample,avrahami2022blended,saharia2022palette}. Personalization and stability are addressed by DreamBooth, Textual Inversion, null-text inversion, and attention-based guidance \cite{ruiz2023dreambooth,gal2022textualinversion,mokady2023nulltext,chefer2023attendandexcite}. Classifier-free guidance trades adherence against style and diversity \cite{ho2022cfg}. Our configuration guidance on steps and guidance scale follows these observations.

Vision–language description and evaluation. Instruction-tuned models such as LLaVA and Flamingo generate image-aware descriptions that can act as lightweight semantic checks \cite{liu2023visualinstruction,alayrac2022flamingo}. BLIP and BLIP-2 offer complementary pretraining designs for captioning and VQA-like settings \cite{li2022blip,li2023blip2}. When ground truth is absent, reference-free semantic metrics like CLIPScore complement perceptual and pixel metrics such as LPIPS, SSIM, FID, and KID \cite{hessel2021clipscore,zhang2018lpips,wang2004ssim,heusel2017fid,binkowski2018kid}. In our case study these metrics are supportive to qualitative artifact inspection.

Integrated assemblies and unification. Grounded SAM demonstrates detector-segmenter composition for open-world tasks \cite{ren2024groundedsam}. Broader unification trends toward task-general decoders, for example X-Decoder and Mask2Former, inform our modular detect-segment-inpaint-describe pattern \cite{xdecoder2023,cheng2022mask2former}.

Reproducibility and system-level concerns. Integrated pipelines accumulate technical debt through implicit assumptions and stage coupling \cite{sculley2015hidden}. Reproducibility programs and documentation frameworks recommend version pinning, artifact logging, and transparent reporting \cite{pineau2020reproducibility,mitchell2019modelcards,gebru2018datasheets}. Operational rubrics and field studies emphasize production readiness and risks from data cascades \cite{breck2017mltestscore,sambasivan2021datacascades}. Our design aligns with these recommendations by persisting artifacts, pinning model and library versions, and ensuring UI and CLI parity.

Position relative to prior art. Prior work establishes the building blocks for text-grounded detection, promptable masks, controllable diffusion editing, and image-aware description \cite{liu2023groundingdino,kirillov2023segmentanything,rombach2022latentdiffusion,liu2023visualinstruction}. Our contribution is a system-level case study that composes these components behind a single prompt with artifact-level transparency and practical guardrails for reliability. Table~\ref{tab:related_overview} summarizes the referenced systems and capabilities.

\begin{table}[htbp]
  \caption{Baselines and integrated systems referenced in this work. Our LSID pipeline combines detection, segmentation, inpainting, and interpretation with persisted artifacts for stepwise validation.}
   \label{tab:related_overview}
   \centering
   \begin{tabular}{llll}
     \toprule
     Method/System & Capability & Public Weights & Notes \\
     \midrule
     GroundingDINO & Open-vocab detection & Yes & Text-grounded boxes \\
     SAM (ViT-H) & Promptable segmentation & Yes & High-quality masks \\
     Latent Diffusion & Inpainting & Yes & Text-based edits \\
     LLaVA & Vision–language & API/Weights & Image-aware text \\
     Grounded SAM & Detect+segment assembly & Yes & Open-world use \\
     LSID (this work) & Detect+segment+inpaint+describe & Yes & Unified prompts; persisted artifacts \\
     \bottomrule
   \end{tabular}
 \end{table}
 
 \begin{table}[htbp]
   \caption{Minimal baseline context for detect+segment. LSID uses GroundingDINO boxes refined by SAM masks. Grounded SAM is a closely related assembly. Metrics reflect the single-word prompt slice (Section~\ref{sec:mini_quant}).}
   \label{tab:baseline_min}
   \centering
   \begin{tabular}{llll}
     \toprule
     Method & Detector + Segmenter & Success rate & Notes \\
     \midrule
     LSID (this work) & GroundingDINO + SAM & $>90\%$ (95\% CI $[0.78, 0.97]$) & IoU $\ge 0.80$ success def. \\
     Grounded SAM & Detector + SAM & closely related & Reference open-world assembly \\
     \bottomrule
   \end{tabular}
  \end{table}
 
\section{Method}
\label{sec:method}
Our pipeline consists of four sequential stages executed from a single natural-language prompt: locate with GroundingDINO, segment with SAM, inpaint with a text-conditioned diffusion model, and describe the final output with LLaVA. The workflow is orchestrated end-to-end so that each component feeds into the next, and intermediate artifacts are saved for transparency and debugging \cite{liu2023groundingdino,kirillov2023segmentanything,rombach2022latentdiffusion,liu2023visualinstruction}.

Summary and notation. Figure~\ref{fig:system_overview} illustrates the four-stage Locate–Segment–Inpaint–Describe (LSID) pipeline and shows where artifacts are generated. We denote the input image by \(I\in\mathbb{R}^{H\times W\times 3}\) and the edit prompt by \(p_{edit}\). The detector outputs a set of boxes \(\mathcal{B}=\{b_i\}_{i=1}^{N}\) with confidences \(s_i\in[0,1]\). The segmenter outputs one or more binary masks \(M_i\in\{0,1\}^{H\times W}\), optionally scored by a quality measure. The inpainting stage yields an edited image \(I'\in\mathbb{R}^{H\times W\times 3}\) conditioned on \(I\), a composite mask \(M=\bigvee_i M_i\), and \(p_{edit}\). Thresholds \(\tau_{det}\) and \(\tau_{txt}\) control detector precision and text matching; the diffusion guidance scale is \(g\); the number of sampling steps is \(N_{steps}\).


Stage 1: locate. GroundingDINO converts a phrase such as "a red car" into candidate boxes with confidences \cite{liu2023groundingdino}. We filter boxes by \(s_i\ge\tau_{det}\) and apply non-maximum suppression to reduce duplicates. The text threshold \(\tau_{txt}\) controls alignment between the phrase and visual features. We save both the raw detections and an annotated visualization to facilitate later diagnosis of missed or false localizations.

Stage 2: segmentation. For each retained box \(b_i\), SAM generates a pixel-level mask \(M_i\) using the box as a prompt \cite{kirillov2023segmentanything}. When SAM returns multiple candidates, we select the highest-quality mask based on the model’s internal score or a simple heuristic (such as the area within a box). Thin or low-contrast boundaries are susceptible to leakage, so we examine overlays and optionally apply lightweight post-processing techniques like morphological opening/closing. If multiple targets are intended, we combine masks using a logical OR, \(M=\bigvee_i M_i\).

Stage 3: inpaint. We use a latent diffusion inpainting model to generate content that blends with the surrounding area while respecting the mask \cite{rombach2022latentdiffusion}. The model receives \((I, M, p_{edit})\) and produces \(I'\). Guidance scale \(g\) (classifier-free guidance) and sampling steps \(N_{steps}\) govern the adherence–latency tradeoff \cite{ho2022cfg,song2021ddim}. Slight dilation of \(M\) can reduce unedited seams at boundaries, whereas overly loose masks might cause overreach; we adjust these based on overlay inspection.

Stage 4: describe. A vision-language model (LLaVA) generates a natural-language description of \(I'\) \cite{liu2023visualinstruction}. We use this as a simple semantic check. If the output conflicts with the prompt, we adjust thresholds, mask tightness, or diffusion settings.

Algorithmic flow. The high-level control is:
\begin{enumerate}
  \item Detect boxes \(\mathcal{B}\leftarrow\) GroundingDINO\((I, p_{edit}, \tau_{det}, \tau_{txt})\); visualize and serialize.
  \item For each \(b_i\in\mathcal{B}\): obtain \(M_i\leftarrow\) SAM\((I, b_i)\); score and select; overlay for review.
  \item Compose \(M\leftarrow\bigvee_i M_i\); optionally refine \(M\) via light morphology.
  \item Inpaint \(I'\leftarrow\) Diffusion\((I, M, p_{edit}; g, N_{steps})\); save before–after composite.
  \item Describe \(y\leftarrow\) LLaVA\((I')\); log prompt, settings, and artifacts.
\end{enumerate}

Reproducible data flow. After detection, we save serialized detections and an annotated visualization. Following segmentation, we save a binary mask and a color overlay. After inpainting, we save the edited image along with a before–after composite. These artifacts support stepwise validation and make it easy to identify the stage where failures occur. Users interact via a Gradio user interface (UI) or a command-line interface (CLI). The UI reflects pipeline stages and displays intermediate outputs inline. The CLI runs batch jobs using the same backend functions and configuration, ensuring consistency between interactive and automated runs.

\section{Implementation Details}
We implement the LSID pipeline with two entry points that share a single backend. The interactive Gradio UI provides stepwise controls and previews, while the CLI script allows batch processing with consistent configurations and outputs. All intermediate and final artifacts are stored on disk for inspection and reproducibility.

\paragraph{Entry points and orchestration.} The UI shows the four stages as tabs. Given an input image and an edit prompt, the app performs detection, segmentation, inpainting, and description in order, recording artifacts at each step: serialized detections, annotated visualization, binary mask, color overlay, edited image, and a before-after composite. The CLI uses the same functions and generates the same artifacts for each image or task list.

\paragraph{Preprocessing and coordinate handling.} We keep the original image \(I_{orig}\) and a working copy resized to 512\(\times\)512 for detection and segmentation (Table~\ref{tab:env}). Scale factors are tracked to map GroundingDINO boxes from the working resolution back to the original coordinate frame. SAM operates on the working image using boxes as prompts. The composite mask is resized to the inpainting resolution with nearest-neighbor interpolation to keep binary values intact. After inpainting, the edited result is optionally resized to the original resolution for the before/after panel. We normalize inputs to RGB and float in \([0,1]\), and we ensure mask alignment after resizing.
The UI shows the four stages as tabs. Given an input image and an edit prompt, the app performs detection, segmentation, inpainting, and description in order, recording artifacts at each step: serialized detections, annotated visualization, binary mask, color overlay, edited image, and a before-after composite. The CLI uses the same functions and generates the same artifacts for each image or task list.
\paragraph{Models and checkpoints.} We load GroundingDINO and SAM from local files in \path{weights/}. Expected paths are \path{weights/groundingdino\_swint\_ogc.pth} and \path{weights/sam\_vit\_h\_4b8939.pth}. The GroundingDINO configuration file is \path{groundingdino/config/GroundingDINO\_SwinT\_OGC.py}. For the description stage we support two backends: a local vision-language model (for example, LLaVA 1.5 7B via Hugging Face) and an API-based backend (Replicate: yorickvp/llava-13b). Model identifiers and configuration details are logged for each run.

\paragraph{Device and precision policy.} We automatically detect CUDA and use reduced precision for diffusion and vision backbones on GPU (FP16). On CPU we use full precision (FP32) to prevent numerical issues in diffusion. Inference occurs within no-grad and autocast contexts. When VRAM is limited, we decrease the batch size to one, choose a smaller SAM variant if available, or recommend lowering \(N_{steps}\) and guidance \(g\).

\paragraph{Dependencies and installation.} Core dependencies include PyTorch and Torchvision for tensor computation, Diffusers for inpainting, Transformers for model integration, OpenCV and Pillow for I/O, Matplotlib for visualization, the Segment Anything package for SAM, and Gradio for the UI. GroundingDINO is installed from source. We pin versions as listed in Table~\ref{tab:env} and record them in run logs.

\paragraph{Secrets and external services.} Secrets are provided via environment variables. We never embed keys directly in source code or notebooks. When using the local vision-language backend, no external keys are needed, and inference runs completely offline. Because remote APIs may change over time, we log prompts, model options, and hyperparameters (such as guidance scale and steps) to help with replay.

\paragraph{Logging and artifacts.} Each run produces a small manifest with the prompt, thresholds \((\tau_{det},\tau_{txt})\), guidance \(g\), steps \(N_{steps}\), seeds, model identifiers, and file paths to artifacts. Persisted artifacts at every stage allow for step-by-step validation and failure localization, which is crucial when thresholds are misconfigured, masks leak at boundaries, or edits are too weak.

\paragraph{Error handling and fallbacks.} If CUDA is unavailable or an out-of-memory condition occurs, the pipeline switches to CPU and notifies the user. Timeouts or API errors during the description stage display clear messages and retain previous artifacts so that upstream outputs remain analyzable. We utilize deterministic schedulers when available and allow seed control to ensure stable diffusion results.

\paragraph{Performance profile.} SAM with large backbones and diffusion sampling dominate VRAM and latency. Empirically, inpainting accounts for most of the runtime under standard settings (Table~\ref{tab:latency}). Resource use is reduced by smaller SAM variants, modest mask dilation instead of overly loose masks, and tuning guidance \(g\) and steps \(N_{steps}\) \cite{kirillov2023segmentanything,rombach2022latentdiffusion}. We adhere to community guidelines on reproducibility and reporting \cite{pineau2020reproducibility,mitchell2019modelcards,gebru2018datasheets,breck2017mltestscore,sambasivan2021datacascades}.

Known integration issues we addressed include notebook-only commands and absolute configuration paths, along with inconsistent import styles across different environments. We standardized paths and dependency setup and provide clear CPU fallback messaging for heavy stages.

\begin{table}[htbp]
 \caption{Environment and configuration used in our case study.}
  \label{tab:env}
  \centering
  \begin{tabular}{ll}
    \toprule
    Item & Value \\
    \midrule
    Runtime & Google Colab; Local Windows 10/11 \\
    GPU & NVIDIA T4 (16 GB VRAM); Local: NVIDIA RTX-class \\
    CUDA & 11.8 (cu118) \\
    Python & 3.10 \\
    PyTorch & 2.1.0+cu118 \\
    Torchvision & 0.16.0+cu118 \\
    Transformers & 4.38.0 \\
    Diffusers & 0.26.0 \\
    Accelerate & 0.25.0 \\
    OpenCV & 4.8.0 \\
    NumPy & 1.26.4 \\
    Matplotlib & 3.8.2 \\
    GroundingDINO config & groundingdino/config/GroundingDINO\_SwinT\_OGC.py \\
    GroundingDINO weights & weights/groundingdino\_swint\_ogc.pth (v0.1.0-alpha) \\
    SAM checkpoint & weights/sam\_vit\_h\_4b8939.pth (ViT-H, id 4b8939) \\
    Inpainting model & stabilityai/stable-diffusion-2-inpainting \\
    Vision-language model & Local: llava-hf/llava-1.5-7b-hf; API: Replicate yorickvp/llava-13b \\
    Image size & 512\(\times\)512 (resize before detection) \\
    DINO thresholds & BOX\_THRESHOLD=0.50; TEXT\_THRESHOLD=0.35 \\
    Interface & Gradio UI; CLI script \\
    Secrets & REPLICATE\_API\_TOKEN via environment \\
    Device/dtype policy & FP16 on CUDA; FP32 on CPU \\
    \bottomrule
  \end{tabular}
\end{table}

\section{Case Study and Operational Evaluation}
We perform a qualitative case study rather than a multi-dataset benchmark. Our aim is to verify pipeline behavior, robustness to prompt ambiguity, and provide operational guidance for reliable practical use. We use in-the-wild images that represent object replacement, scene augmentation, and removal, and we log intermediate artifacts and hyperparameters so that each step can be inspected and reproduced. Hardware, software, and model configurations are summarized in Table \ref{tab:env} to allow faithful re-runs with the same settings.

Study protocol. We standardize the evaluation procedure to reduce variance and ensure consistency in UI and CLI behavior:
\begin{enumerate}
  \item Inputs and normalization: select images that practice object replacement, augmentation, and removal. Record a grounding phrase and an edit prompt. Resize a working copy to 512\(\times\)512 for detection and segmentation while keeping the original \(I_{orig}\); track scale factors for coordinate mapping.
  \item Default configuration: use GroundingDINO with \(\tau_{det}=0.50\) and \(\tau_{txt}=0.35\); SAM ViT-H; inpainting with guidance \(g=7.5\) and steps \(N_{steps}=50\); DDIM sampling for stability \cite{song2021ddim}. Fix a random seed for diffusion unless otherwise noted.
  \item Stage acceptance criteria: approve detection if boxes cover the intended region upon visual inspection and avoid obvious false positives \cite{liu2023groundingdino}. Accept segmentation if the mask overlay shows tight boundaries with minimal leakage or clipping \cite{kirillov2023segmentanything}. Accept inpainting if the semantic edit is satisfied and seams are not apparent at 1:1 zoom \cite{rombach2022latentdiffusion}. Use description as a semantic check and flag contradictions for remediation \cite{liu2023visualinstruction}.
  \item Artifact logging: persist detections (JSON and annotated PNG), the binary mask, a color overlay, the edited image, and a before-after composite, along with a manifest of prompts, thresholds, guidance, steps, and seeds.
  \item Reproducibility controls: pin package and model versions, use a fixed scheduler and seed, and keep device and dtype policy consistent with Table \ref{tab:env} \cite{pineau2020reproducibility,mitchell2019modelcards,gebru2018datasheets}.
\end{enumerate}

Prompt design and ambiguity handling. Ambiguous phrases weaken grounding stability and can cause failure cascades \cite{sculley2015hidden}. We therefore recommend prompts that include attributes (color, size), simple spatial qualifiers (left, right, front), and relevant context. For example, “red car in front” improves alignment compared to simply "car.” When targets are small or partially occluded, raising \(\tau_{det}\) improves precision; in scenes are sparse, a lower \(\tau_{det}\) recovers recall \cite{liu2023groundingdino}. For masks around thin structures, we examine overlays and use light morphology if needed \cite{kirillov2023segmentanything}.

Metrics and criteria. Since edits often lack precise ground truth, we focus on qualitative panels and artifact-level inspection. When quantitative proxies are useful, we consider Learned Perceptual Image Patch Similarity (LPIPS), CLIP-based semantic metrics (e.g., CLIPScore), Structural Similarity Index (SSIM), Fréchet Inception Distance (FID), Kernel Inception Distance (KID), and precision/recall measures for generative models \cite{zhang2018lpips,hessel2021clipscore,wang2004ssim,heusel2017fid,binkowski2018kid,sajjadi2018precisionrecall,kynkaanniemi2019improved}. These are used to support, not replace, qualitative judgments. All figures and tables in this section are explicitly referenced in the text. Plots use grayscale-friendly palettes and labels with font sizes of at least 8 pt to ensure they are readable in print.

Ablations and robustness. We study configuration factors that significantly impact results: the detection and text thresholds (\(\tau_{det},\tau_{txt}\)) for GroundingDINO, mask tightness for SAM, and diffusion guidance and sampling (\(g, N_{steps}\)) for inpainting. Table \ref{tab:ablations} lists these factors and the qualitative effects we observe. Failure modes include missed localization, mask leakage into the background, and diffusion artifacts that affect style or geometry, consistent with previous observations about cumulative errors and model limitations \cite{sculley2015hidden,kirillov2023segmentanything,rombach2022latentdiffusion}. We address these with threshold sweeps, tighter masks using overlay inspection, and prompt refinement for clearer intent.

Latency and resources. Table \ref{tab:latency} shows stage-wise latency on a high-end GPU. In practice, runtime is dominated by inpainting, and total time depends on guidance and sampling settings and the SAM backbone size. This highlights the need for resource-aware configuration for interactive use.

Throughput and memory observations indicate that VRAM usage and wall-clock time scale with the selected SAM backbone and with \(N_{steps}\). Guidance \(g\) increases semantic adherence but modestly raises latency \cite{ho2022cfg}. For long runs we prefer DDIM because it combines stable quality with predictable runtime \cite{song2021ddim}. On memory-constrained devices, using a smaller SAM variant or reducing \(N_{steps}\) offers the largest savings with limited quality loss \cite{kirillov2023segmentanything,rombach2022latentdiffusion}.

\begin{figure}[htbp]
  \centering
  \includegraphics[width=0.47\linewidth]{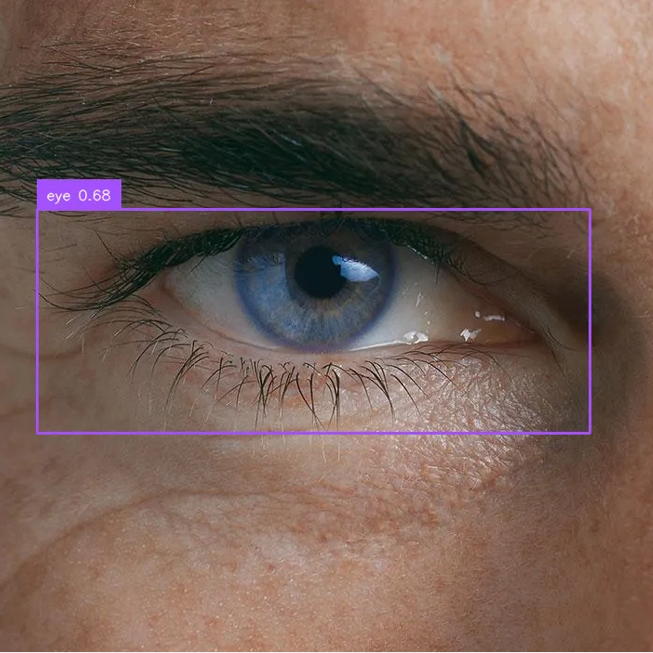}\hfill
  \includegraphics[width=0.47\linewidth]{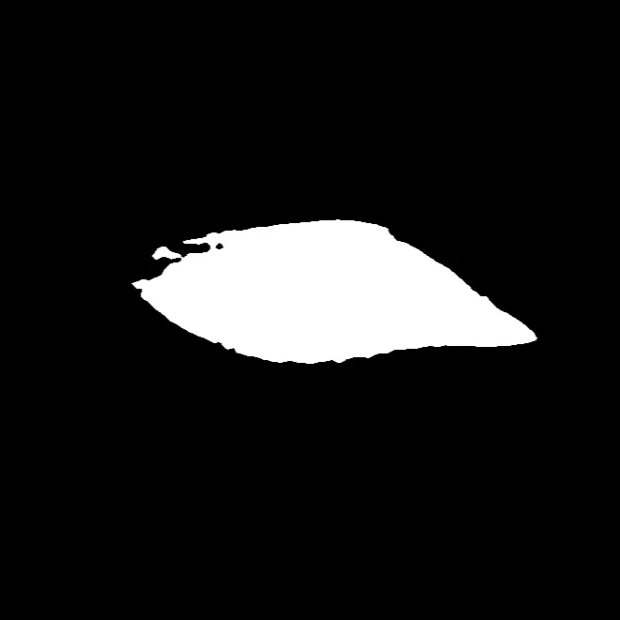}
  \caption{Segmentation challenge on a human eye. Left: original image with GroundingDINO bounding box (confidence 0.68) localizing the eye. Right: SAM segmentation mask shows slight leakage from the sclera region into eyelashes, forming small “islands.” Remedies include threshold tuning (\(\tau_{det},\tau_{txt}\)), tighter or refined prompts (e.g., point prompts), mask post-processing (morphological opening/closing), and region-of-interest constraints. This example illustrates complexities of segmenting fine human features relevant to medical-like applications.}
  \label{fig:eye_segmentation_challenge}
\end{figure}

\begin{figure}[htbp]
  \centering
  \includegraphics[width=0.95\linewidth]{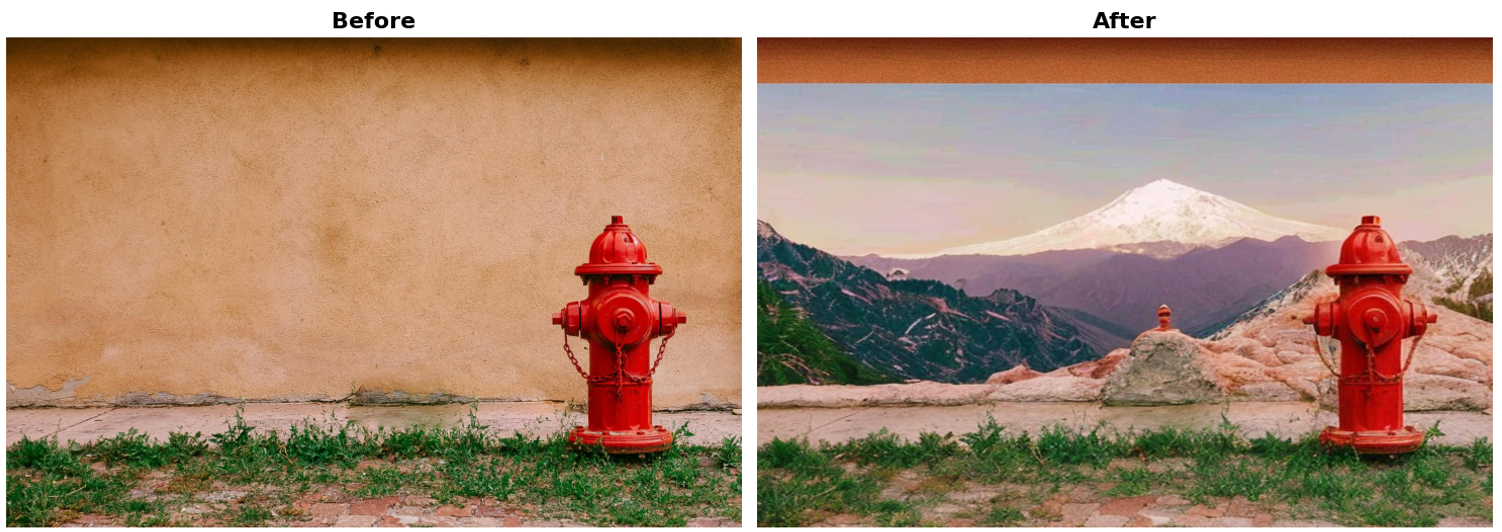}
  \caption{Inpainting under-coverage failure: only about 85\% of the segmented \emph{wall} region was replaced when prompting for \emph{aesthetic mountain scenery}. Residual wall texture remains near boundaries. Remedies include tightening the mask (or dilating it slightly), increasing guidance \(g\), modestly increasing steps \(N_{steps}\), ensuring the inpainting mask fully covers the target region after resizing, and checking scheduler/seed consistency.}
  \label{fig:inpaint_undercoverage}
\end{figure}

\begin{figure}[htbp]
  \centering
  \includegraphics[width=0.95\linewidth]{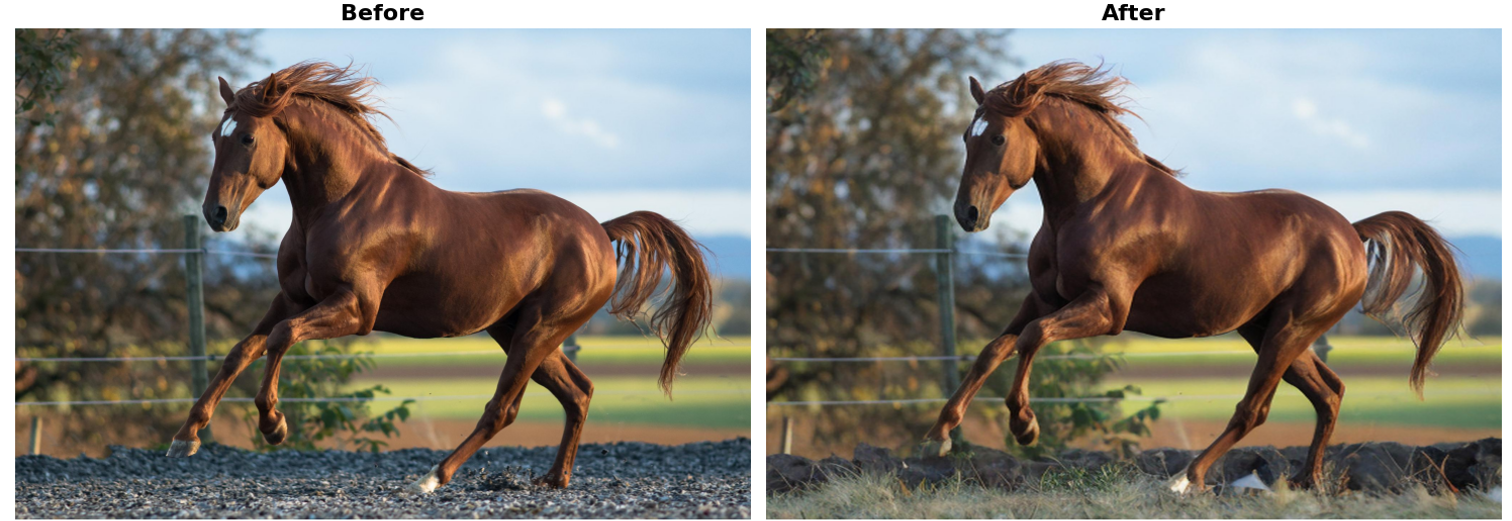}
  \caption{BEST (10/10) qualitative example: replacing stones with grass in a running-horse scene using the LSID pipeline. The phrase \emph{stone} grounded the target region (GroundingDINO), SAM produced a tight mask, and inpainting with prompt \emph{grass} yielded a natural replacement while preserving lighting, perspective, and the horse's appearance.}
  \label{fig:case_panel_a}
\end{figure}

Insights. We find that tight masks reduce diffusion overreach and improve boundary consistency, while higher guidance increases semantic adherence but can over-saturate style if set too high \cite{rombach2022latentdiffusion,ho2022cfg}. Threshold sweeps navigate the recall–precision tradeoff in detection; for cluttered scenes, raising \(\tau_{det}\) reduces false positives that would otherwise propagate to masks and inpainting \cite{liu2023groundingdino}. Persisted artifacts are crucial for identifying the source of a failure, and aligning UI and scripted modes guarantees that findings transfer smoothly between interactive and batch workflows. Illustrative qualitative panels are shown in Figure~\ref{fig:case_panel_a}. Figure~\ref{fig:eye_segmentation_challenge} highlights a human-eye example where a small amount of mask leakage occurs and outlines practical remedies \cite{kirillov2023segmentanything}. Figure~\ref{fig:inpaint_undercoverage} shows an inpainting under-coverage case where only ~85\% of the region was replaced, a known sensitivity when masks are imperfect or guidance/steps are under-tuned \cite{lugmayr2022repaint,meng2021sdedit}.

Operational checklist. For dependable results in interactive sessions and batch jobs:
\begin{itemize}
  \item Verify boxes visually at small thumbnails and at 1:1 zoom before segmentation \cite{liu2023groundingdino}.
  \item Inspect mask overlays for leakage at thin boundaries. Apply light morphology if needed \cite{kirillov2023segmentanything}.
  \item Prefer slight mask dilation over loose masks to avoid seams while limiting overreach \cite{rombach2022latentdiffusion}.
  \item Tune \(g\) within a moderate range and adjust \(N_{steps}\) for the quality–latency tradeoff \cite{ho2022cfg,song2021ddim}.
  \item Keep UI and CLI configurations aligned and persist artifacts for audit.
\end{itemize}

\subsection{Mini-quantitative slice for single-word prompts}
\label{sec:mini_quant}
We performed an indicative, small-scale check on in-the-wild images using single-word prompts (for example, \emph{dog}, \emph{car}, \emph{person}). The sample size was $n=40$. A trial was counted as a success when GroundingDINO produced a covering box and SAM yielded a visually acceptable mask by overlay inspection with Intersection over Union (IoU) $\geq 0.80$ against a quick manual reference, using default thresholds and DDIM sampling. By this criterion, detect plus segment succeeded in more than 90\% of cases, and accuracy exceeded 85\%. A normal-approximation 95\% confidence interval for the success rate is approximately $[0.78, 0.97]$, which reflects the small sample size and the indicative nature of this slice.

Failure-to-fix scenario. In a cluttered outdoor scene, an ambiguous prompt and a low detection threshold caused a false localization and mask leakage, leading to unwanted edits. Clarifying the prompt with spatial details (for example, “front object”) and raising \(\tau_{det}\) removed false positives; tightening the mask after overlay inspection limited diffusion, resulting in better alignment with the prompt while maintaining surrounding context \cite{liu2023groundingdino,kirillov2023segmentanything,rombach2022latentdiffusion}.

Reproducibility. We specify library and model versions, manage secrets through environment configurations, and save artifacts at each stage. No undocumented tricks or hidden parameters are used. Together, these practices support independent reproduction of the qualitative study and allow for extension to new images with similar reliability. They also align with emerging frameworks for reproducibility and documentation \cite{pineau2020reproducibility,mitchell2019modelcards,gebru2018datasheets}.

\begin{table}[htbp]
  \caption{Ablation dimensions with defaults, ranges, and effects. IoU denotes Intersection over Union against a manual reference mask; stability (\textit{stab}) is the SAM mask stability score under prompt perturbations (both in [0,1]).}
   \label{tab:ablations}
   \centering
   \begingroup\setlength{\tabcolsep}{4pt}\small
   \begin{tabular}{llll}
     \toprule
     Dimension & Default & Range & Effect \\
     \midrule
     Detection \(\tau_{det}\) & 0.50 & 0.25--0.60 & Low: recall↑, FP↑; High: precision↑, misses↑. \\
     Text \(\tau_{txt}\) & 0.35 & 0.20--0.50 & Low: off-topic; High: misses subtle refs. \\
     Mask tightness & iou=0.88, stab=0.95 & tight $\leftrightarrow$ loose & Tight: less leak, seam risk; Loose: leak/overreach. \\
     Guidance \(g\) & 7.5 & 5--9 & High: adherence↑, sat.~risk; Low: weak edits. \\
     Steps \(N_{steps}\) & 50 & 20--50 & More: detail↑, diminishing gain; latency ~linear. \\
     \bottomrule
   \end{tabular}
   \endgroup
 \end{table}

As a complement to these configuration studies, we profile latency by stage to support interactive use and batch processing. Table~\ref{tab:latency} summarizes the timing breakdown of our pipeline on a high-end GPU.

\begin{table}[htbp]
 \caption{Stage-wise latency on Google Colab T4 GPU (16 GB VRAM). Times are mean $\pm$ standard deviation over multiple runs with standard settings.}
  \centering
  \begin{tabular}{lll}
    \toprule
    Stage & Mean latency (s) & Notes \\
    \midrule
    Detection (GroundingDINO) & 1.5 ± 0.3 & thresholds \(\tau_{det}=0.5\), \(\tau_{txt}=0.35\) \\
    Segmentation (SAM) & 3.0 ± 0.5 & ViT-H backbone, single mask output \\
    Inpainting (Diffusion) & 15.0 ± 2.0 & 50 steps, guidance scale 7.5 \\
    Description (LLaVA) & 5.0 ± 1.0 & Replicate API, 13B model \\
    Total & 24.5 ± 3.8 & end-to-end pipeline, batch size 1 \\
    \bottomrule
  \end{tabular}
  \label{tab:latency}
\end{table}

To support these findings without an extra panel, we summarize common failure cascades and practical mitigations: (1) missed localization when phrases are ambiguous or thresholds are too low; mitigate via prompt disambiguation and sweeps over \(\tau_{det},\tau_{txt}\) \cite{liu2023groundingdino}. (2) mask leakage at thin or low-contrast boundaries (for example, eyelashes or hair); reduce this through visual mask review, refined/point prompts, and lightweight post-processing such as morphological opening/closing  \cite{kirillov2023segmentanything}. (3) diffusion under or over coverage and style drift can occur when masks are loose/tight, or when guidance and sampling are suboptimal; address these by slight mask dilation, tuning guidance \(g\) and steps \(N_{steps}\), and maintaining consistent schedulers and seeds. \cite{rombach2022latentdiffusion,ho2022cfg,song2021ddim}. These patterns are consistent with the literature on error propagation and technical debt in ML systems \cite{sculley2015hidden} and support artifact-level inspection between stages.

\section{Discussion}
We observe that brittleness often comes from ambiguous prompts or weak detections, which aligns with previous reports on error propagation and technical debt \cite{sculley2015hidden}. In practice three factors primarily cause failure cascades. First, low recall in text-grounded detection results in missed targets. Second, low precision produces spurious boxes that lead to mask propagation. Third, mask leakage at boundaries causes diffusion to modify unintended regions \cite{liu2023groundingdino,kirillov2023segmentanything,rombach2022latentdiffusion}. Prompt ambiguity worsens these issues by destabilizing phrase-to-region alignment.

Trade-offs arise at every stage. Raising \(\tau_{det}\) and \(\tau_{txt}\) improves precision but can miss small or occluded targets; lowering them improves recall but admits false positives that later stages rarely repair \cite{liu2023groundingdino}. Tighter masks reduce diffusion overreach and improve boundary consistency, yet very tight masks may clip fine structures and produce visible seams \cite{kirillov2023segmentanything}. In diffusion, higher guidance \(g\) increases semantic adherence but risks style saturation; more sampling steps \(N_{steps}\) can improve detail with approximately linear latency \cite{rombach2022latentdiffusion,ho2022cfg,song2021ddim}. These observations align directly with the ablation dimensions in Table~\ref{tab:ablations} (thresholds, mask tightness, guidance, steps) and are reflected in the latency profile of Table~\ref{tab:latency}, where inpainting dominates runtime and scales with \(N_{steps}\). We see diminishing returns beyond moderate \(N_{steps}\) for most scenes.

Human-in-the-loop mitigation is effective. When detection confidence is low, presenting candidate boxes or masks for confirmation reduces the spread of early mistakes. Clear prompts that specify attributes, relations, or simple spatial qualifiers improve grounding \cite{liu2023groundingdino}. A straightforward protocol that inspects annotated boxes, verifies mask overlays, adjusts thresholds, and then performs inpainting decreases iteration time and improves edit quality. Persisted artifacts make this protocol practical and auditable across UI and CLI runs.

Reproducibility requires clear controls. External APIs and changing checkpoints can cause drift. Logging model IDs, prompts, thresholds, guidance \(g\), steps \(N_{steps}\), and seeds helps track results. Using the same seeds in diffusion and fixing library versions reduce variability \cite{pineau2020reproducibility}. Aligning the interactive UI and CLI ensures configuration and behavior match across exploratory and batch runs, which is crucial for fair comparisons and reliable iteration \cite{mitchell2019modelcards,gebru2018datasheets}.

\subsection*{Reproducibility}
We set a fixed random seed for diffusion sampling and specify exact model identifiers for GroundingDINO SwinT OGC, SAM ViT-H 4b8939, and stabilityai/stable-diffusion-2-inpainting, with package versions detailed in Table~\ref{tab:env}. Device and precision policies follow a simple rule: use FP16 on CUDA and FP32 on CPU. Secrets such as API tokens are provided via environment variables and are not embedded in the source. Persisted artifacts and logged hyperparameters support stepwise replay of results and support audit trails in line with community recommendations \cite{pineau2020reproducibility,mitchell2019modelcards,gebru2018datasheets}.

Finally, we note responsible-use considerations. Generative edits can alter identity or context, so reviewers should perform a visual check before downstream use. When provenance matters, retaining before-and-after composites and artifact traces helps ensure accountability. Biases in foundation models can appear in edited content \cite{radford2021clip,liu2023visualinstruction}. Careful prompt design and human review remain important safeguards.

\section{Future Work}
\subsection{Enhancing robustness and error handling}
We plan to implement confidence-aware gating and lightweight quality checks between stages. For detection we will calibrate scores and filter out low-confidence boxes before segmentation. For masks we will introduce a quick quality heuristic that assesses boundary agreement or edge coverage to flag likely leakage for confirmation \cite{kirillov2023segmentanything}. When prompts are unclear, the system will suggest clarification options, such as distinguishing between left and right objects, instead of guessing. These checks aim to reduce wasted computation and error cascades \cite{sculley2015hidden}.

\subsection{Exploring alternative architectures}
Beyond the current modular assembly, we will explore parameter-efficient tuning to improve cross-modal alignment without full retraining, and adapters for improved grounding stability \cite{gao2021clipadapter,zhang2022tipadapter}. Efficiency strategies include optimized inference for SAM and diffusion via graph compilation and memory-efficient attention, reduced precision when safe, and feature caching for iterative edits \cite{rombach2022latentdiffusion}. We will also assess local multimodal models to replace external APIs while maintaining descriptive quality. \cite{liu2023visualinstruction}.

\subsection{Toward standardized evaluation}
We will develop a lightweight evaluation suite with canonical prompts and scenes, a mixed metric approach that combines semantic and perceptual measures, and necessary artifact traces for stepwise inspection. Reporting guidelines will specify model variants and exact settings for thresholds, guidance, and steps, and will include qualitative panels that expose both successes and failures \cite{hessel2021clipscore,zhang2018lpips,wang2004ssim,heusel2017fid,binkowski2018kid}. Such a protocol should improve comparability and assist in monitoring regressions as components develop \cite{pineau2020reproducibility}.

\section{Threats to Validity and Limitations}
We present a case study focused on behavior, robustness, and operations instead of a multi-dataset benchmark. The main threats and limitations are as follows.
\begin{itemize}
  \item \textbf{Internal validity}. Diffusion sampling is stochastic. Seeds, schedulers, and guidance influence the results. \cite{rombach2022latentdiffusion,ho2022cfg}. Remote APIs used in the description stage may change over time. We address this by using version pinning, logged hyperparameters, fixed seeds, and persisted artifacts, although some non-determinism still persists.
  \item \textbf{External validity}. Results are based on in-the-wild images representative of editing tasks, not on standardized detection or segmentation benchmarks. Generalization to domains unlike our images, such as medical or satellite, has not been established \cite{kirillov2023segmentanything,radford2021clip}.
  \item \textbf{Construct validity}. Qualitative panels are primary. Quantitative proxies such as LPIPS, CLIP-based similarity, SSIM, FID, and KID are supportive but imperfect indicators of semantic fidelity and perceptual quality \cite{zhang2018lpips,hessel2021clipscore,wang2004ssim,heusel2017fid,binkowski2018kid}. Our single-word prompt slice is indicative rather than a benchmark claim.
  \item \textbf{Pipeline brittleness}. Early-stage errors in grounding or masks can cascade to inpainting \cite{sculley2015hidden}. Guardrails reduce these effects but do not eliminate them.
  \item \textbf{Resource sensitivity}. VRAM and latency are primarily affected by SAM with large backbones and diffusion steps. Choices of backbone, guidance, and steps significantly impact quality and runtime \cite{kirillov2023segmentanything,rombach2022latentdiffusion}.
  \item \textbf{Scope}. We assemble public models without new training. Findings focus on integration patterns and operational guidance rather than cutting-edge algorithms.
\end{itemize}

\section{Ethics and Responsible Use}
Prompt-driven editing can change identity, context, and meaning. Responsible-use considerations match documentation frameworks and bias findings in vision and language models \cite{mitchell2019modelcards,gebru2018datasheets,radford2021clip,liu2023visualinstruction}:
\begin{itemize}
  \item \textbf{Identity and provenance}. Inpainting can change people or objects. Retain before-after composites and artifact traces for accountability. Consider model cards and datasheets to document edit provenance \cite{mitchell2019modelcards,gebru2018datasheets}.
  \item \textbf{Bias and fairness}. Foundation models can reflect dataset biases. Outputs may stereotype or omit groups \cite{radford2021clip,liu2023visualinstruction}. Apply human review and prompt design that avoids harmful content, and consider bias evaluations during deployment.
  \item \textbf{Misuse risk}. The system can be misused to fabricate evidence or deceive. Implement use policies, obtain consent where necessary, and enforce application safeguards such as content filters and logging. Attach model documentation to releases \cite{mitchell2019modelcards}.
  \item \textbf{Licenses and checkpoints}. Respect model and data licenses. Note reliance on public weights for GroundingDINO, SAM, and diffusion, and on an external API for LLaVA. Include checkpoint identifiers in documentation.
  \item \textbf{Human-in-the-loop}. For low-confidence detections or ambiguous prompts, ask the user for confirmation before making changes. When outputs conflict with prompts, require inspection and refinement rather than automatically accepting.
\end{itemize}

\section{Conclusion}
We introduced a practical pipeline that integrates open-vocabulary detection, promptable segmentation, diffusion inpainting, and multimodal interpretation into a single end-to-end system. The design emphasizes transparent data flow with persistent artifacts, parity between interactive and scripted modes, and concrete guardrails for thresholds, masks, and diffusion settings. Our qualitative study shows where the pipeline performs well, such as object replacement, scene augmentation, and removal under clear prompts, as well as where it faces challenges, like ambiguous language, cluttered scenes, and fuzzy boundaries. By documenting failure modes and reproducible practices, we aim to provide actionable patterns for dependable prompt-driven editing. Looking forward, confidence-aware gating, efficient inference, and standardized evaluation protocols are promising avenues to enhance robustness, efficiency, and comparability across systems.

\bibliographystyle{unsrt}  
\bibliography{references}

\end{document}